\documentclass[sigconf]{acmart}

\usepackage{soul}
\usepackage{bookmark}
\usepackage{amsfonts}
\usepackage{enumitem}

\usepackage[ruled,vlined,linesnumbered]{algorithm2e}
\usepackage{mathtools}
\usepackage{subcaption}
\setlist[itemize]{leftmargin=*}
\setlist[enumerate]{leftmargin=*}

\graphicspath{ {./figures/} }

\AtBeginDocument{%
  \providecommand\BibTeX{{%
    \normalfont B\kern-0.5em{\scshape i\kern-0.25em b}\kern-0.8em\TeX}}}

\setcopyright{acmcopyright}
\copyrightyear{2022}
\acmYear{2022}
\acmDOI{}

\acmConference[AutoML, KDD '22]{KDD '22}{August 14--17, 2022}{Washington, DC, USA}
\acmBooktitle{The Sixth International Workshop on 
Automation in Machine Learning}
\acmPrice{}
\acmISBN{}%978-1-4503-XXXX-X/18/06}

\begin{document}

\settopmatter{printfolios=true}
\title{Lightweight Automated Feature Monitoring for Data Streams}

\author{João Conde}
\email{joao.conde@feedzai.com}
\authornote{Work developed while employed at Feedzai.}
\affiliation{%
%  \institution{Feedzai}
  \country{Feedzai}
}
\author{Ricardo Moreira}
\email{ricardo.moreira@feedzai.com}
\affiliation{%
%  \institution{Feedzai}
  \country{Feedzai}
}
\author{João Torres}
\email{joao.torres@feedzai.com}
\affiliation{%
%  \institution{Feedzai}
  \country{Feedzai}
}

\author{Pedro Cardoso}
\authornotemark[1]
\email{pedro.cardoso@feedzai.com}
\affiliation{%
%  \institution{Feedzai}
  \country{Feedzai}
}

\author{Hugo Ferreira}
\email{hugo.ferreira@feedzai.com}
\affiliation{%
%  \institution{Feedzai}
  \country{Feedzai}
}

\author{Marco O. P. Sampaio}
\email{marco.sampaio@feedzai.com}
\affiliation{%
%  \institution{Feedzai}
   \country{Feedzai}
}

\author{Jo\~ao Tiago Ascens\~ao}
\email{joao.ascensao@feedzai.com}
\affiliation{%
%  \institution{Feedzai}
   \country{Feedzai}
}

\author{Pedro Bizarro}
\email{pedro.bizarro@feedzai.com}
\affiliation{%
%  \institution{Feedzai}
   \country{Feedzai}
}

\renewcommand{\shortauthors}{Conde et al.}

\begin{abstract}
Monitoring the behavior of automated real-time stream processing systems has become one of the most relevant problems in real world applications. Such systems have grown in complexity relying heavily on high dimensional input data, and data hungry Machine Learning (ML) algorithms. We propose a flexible system, Feature Monitoring (FM), that detects data drifts in such data sets, with a small and constant memory footprint and a small computational cost in streaming applications. The method is based on a multi-variate statistical test and is data driven by design (full reference distributions are estimated from the data). It monitors all features that are used by the system, while providing an interpretable features ranking whenever an alarm occurs (to aid in root cause analysis). The computational and memory lightness of the system results from the use of Exponential Moving Histograms. In our experimental study, we analyze the system's behavior with its parameters and, more importantly, show examples where it detects problems that are not directly related to a single feature. This illustrates how FM eliminates the need to add custom signals to detect specific types of problems and that monitoring the available space of features is often enough.
\end{abstract}



%datasets, neural networks, gaze detection, text tagging}


%\tableofcontents

\begin{teaserfigure}
  \includegraphics[width=\textwidth, clip=true]{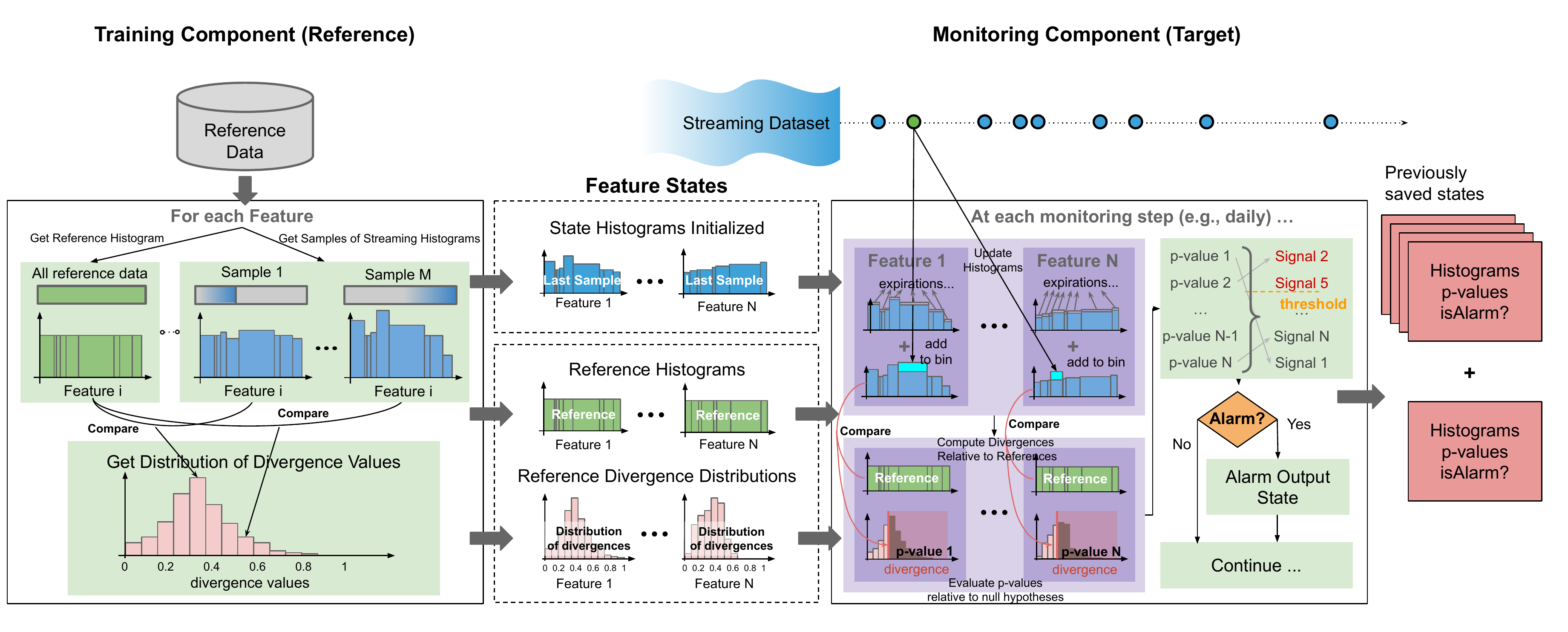}
  \caption{Feature Monitoring System Overview}
  \label{fig:system-architecture}
\end{teaserfigure}

\maketitle

\section{Introduction}\label{sec:introduction}

Real-time stream processing systems have become ubiquitous in recent times.
The way they are set up often implicitly assumes that future data flowing through the system will always follow the same distribution.
Thus, even though the system may initially perform well, over time, due to data drift, a static configuration may result in performance deterioration, requiring reconfiguration over time.

In the financial services industry, data drift can occur due to expected (e.g., seasonality) or unexpected factors, such as market shifts, new buying patterns introduced by disruptive technologies, or some technical issues that might corrupt the observed data. 
Furthermore, the drift can be gradual or sudden, \cite{lu2018learning},  and it may occur in specific features of the data or collectively on all of them.
Machine Learning (ML) models responsible for predictive tasks can make absurd decisions extrapolating from the short-sighted training set observations, impacting businesses and system users.

Accurate and timely detection of data drift allows for early mitigation of its effects.
This challenging task is made easier when labels arrive instantaneously and real-time evaluations are possible. 
However, in many domains, such as fraud detection~\cite{DBLP:journals/corr/abs-2107-01979},
labels may not be available for several weeks.
In this scenario, unsupervised methods are necessary to detect data drift in a timely manner, accelerate corrective action and minimize business costs.
Furthermore, in many streaming scenarios, data is generated at a very high rate. Thus, if real-time monitoring is required, lightweight solutions with low computational, storage and memory costs are essential.  

In this article, we present a system to automate the detection of data drifts based on monitoring the distribution of data features, which we refer to as \textit{Feature Monitoring}.
Our solution guarantees a low memory footprint, using histograms to summarise distributions, as well as low computational costs (hence low real-time latencies) using methods that support recursive updates.
The drift detection and alarming is based on a multivariate data-driven statistical test using a reference data period. Thus, our method does not make strong assumptions on the data distributions or their drifts. 

Besides producing alarms, the system outputs continuous signals that can be used to visualize the global state of the data being monitored. The various signals are mapped directly to interpretable features that provide an explanation via a ranking of the features that are responsible for the alarm. The statistical test deals with the problem of multiplicity correcting for multiple hypotheses testing. 

Overall, our system differs from the existing literature (see also the related work in Sect.~\ref{sec:related-work}) in that it combines several characteristics in one system, namely, (i) it provides a computational and memory efficient method for streaming data, (ii) it is intrinsically multi-variate and multi-signal (monitoring a multi-dimensional feature space), and (iii) it provides an explanation to point the user to the source of the problem. One crucial ingredient in our data-driven approach is that we sample various periods to define the reference distribution of diverging behavior. This effectively allows for a large tolerance to temporal changes as observed in the training, which in turn avoids an over alarming system.

In sum, our main contributions are:
\begin{itemize}
    \item A data-driven method to build a reference representation of a data period and its fluctuations (Sect.~\ref{sec:batch-phase}).
    \item A computational and memory efficient monitoring procedure
    that defines alarms and provides an explanation on which features are important to explain the alarms (Sect.~\ref{sec:monitoring-component}).
    \item An empirical study of the method for several real world datasets, including a public dataset, in the fraud detection domain. This includes studies of: parameters sensitivity (Sect.~\ref{subsec:parameter_effects}), detection of real world drifts (Sect.~\ref{subsec:merchant_signal}), and of injected drifts (Sect.~\ref{subsec:public_drift}). 
\end{itemize}

\section{Methods}
\label{sec:methods}

We propose a system to monitor the distribution of features (categorical or numerical) of a dataset.
We develop an efficient method that can both be used in a streaming production environment for monitoring, as well as in offline data exploration, which we denote as \textit{Feature Monitoring} (FM).
The system is application agnostic, however a common use case is when features are used by a ML model for predictive tasks or in decision systems based on rules. 

In Fig.~\ref{fig:system-architecture}, we present a schematic overview of the system. The two main components are summarized in the box diagrams on the left (\emph{Training}) and on the right (\emph{Monitoring}), connected by the Feature States computed in \emph{Training} (dashed line blocks). In the next sections we describe each component in detail.

\subsection{Train References Component}
\label{sec:batch-phase}

In this section, we describe the \emph{Training} component (left block diagram of Fig.~\ref{fig:system-architecture}). 
Its purpose is to estimate reference distributions for each of the features being monitored. This is based on a fixed \textit{Reference Data} source containing several events (grey cylinder).
Typically, the reference should comprise an extended period of several weeks or months of data, depending on the stream's event rates. This can correspond, for instance, to a ML model training period.
In the next sub-sections, we describe each step of this component.

\subsubsection{Reference Histograms}

Given a set of features to monitor, the first step of the method is to build an overall reference histogram $H_{R,f}$ for each feature $f=1,\ldots,N$ to characterize the training data distribution in the reference period, $X_T$, (top left green panel in the \emph{Training} block of Fig.~\ref{fig:system-architecture}). $X_T$ should be a representative sample of the whole training period. For each $H_{R,f}$ we use a set of equal-frequency, i.e., quantile, bins (see 
%line \ref{batch_buildRef} of algorithm %\ref{batch_algo}, and also the 
leftmost green block in Fig.~\ref{fig:system-architecture}) to cover the densest regions of the distribution with a larger number of bins. Our histograms are built using $b + 3$ bins where $b$ bins are used to cover all existing values in the distribution of the reference data, and 3 additional bins are added:  $\left]-\infty,\,\text{bin}(1)_{\min}\right]$ (leftmost), $\big]\text{bin}(b)_{\max} ,\,+\infty\big[$ (next to rightmost) and $\left[\mathtt{NaN}\right]$ (rightmost). The semi-infinite bins cover regions not observed in the reference dataset. In contrast, the $\mathtt{NaN}$ bin is necessary for instances with invalid feature values (e.g., an empty value due to a feature collection problem). This way we ensure that the histograms always have full support for any possible value. 

\subsubsection{Sampling of Time Steps}
\label{subsubsec:sampling-of-timestamps}
The second step in building the reference is to randomly sample observation time steps in the reference period. This is motivated by the goal of monitoring feature values in an online production environment, usually covering timescales that are shorter than the reference period.
For example, assume that the reference data corresponds to six months used to train an ML model. We may wish to monitor the behavior of the features in one week periods after the model was deployed in streaming. 

Our main assumption is that the reference period defines the expected distribution of data shifts. Hence, at each time step, we want to monitor the distribution of a feature's values by comparing its distribution in the original reference (or training) period with the distribution in shorter monitoring periods.  
To perform this comparison, we use a divergence measure (details in Section~\ref{subsubsec:streaming-histograms}) to obtain the histogram of divergence values, $H_{D,f}$ (see the bottom green block in Fig.~\ref{fig:system-architecture}) by computing the divergences of $H_f$ histograms for all sampled time steps with the reference one $H_{R,f}$.

Computing a feature histogram and subsequent divergence value for each event can be very heavy for realistic datasets with millions of events per day.
To bypass this problem, we resort to random sampling to obtain a smaller number of time steps at which to compute the divergence values.
%The main question in this procedure is: What is the minimum number of samples that we should use? 
For this sampling to be representative, we need a minimum number of samples, $M$, which we estimate next.

Note that the proposed system focuses on monitoring events for which the divergence values become large, i.e., in the higher quantiles of $H_{D,f}$.
This requirement means that a good estimate of the upper tails of all the distributions is needed.
Our monitoring approach (described in more detail in later sections) will be to perform a statistical test under a multiple null hypothesis that the divergence values observed 
in \textit{Streaming} follow the same distributions as the $H_{D,f}$. 
The multiple hypothesis test requires setting a global significance level (family-wise error rate) $\bar{\alpha}$ for its composite null hypothesis, which corresponds to the $p$-value of rejecting it due to random fluctuations.
This usually results in a much stricter significance level applied to at least some of the individual hypotheses, since the probability that at least one of the tests fails by chance grows with the number of tests. 

To obtain a conservative upper bound on the critical level for any feature, we first refer to the Bonferroni correction, which is valid even if the hypotheses are dependent.
Therefore, if any of the $N$ individual hypothesis fails the test at a level $\alpha=\bar{\alpha}/N$ then the multiple hypothesis fails at a level $\bar{\alpha}$.
We aim to ensure that our divergence histograms have enough points to estimate the $\alpha$-upper-tail appropriately based on this conservative bound.
If the number of samples produced to represent $H_{D,f}$ is $M$, then the probability, $p_0$, that none of those samples falls on the tail (assuming independent samples) is $p_0 = (1-\alpha)^M$.
Furthermore, because we are building $N$ histograms, we need to limit the probability that any histograms are missing samples in the tail of the distribution. 
The probability, $\gamma$, that one or more histograms  miss samples on the tail is related to the probability that none of them miss samples on the tail\footnote{For this estimate, for simplicity, we now assume independence between the features.}:
\begin{equation}
\gamma = 1-(1-p_0)^N = 1-\left(1-(1-\alpha)^M\right)^N\; .
\end{equation}
This limits the probability of having one or more ``tail''-incomplete histograms. Inverting this formula and replacing $\alpha$ by $\bar{\alpha}/N$ we obtain that the minimum number of samples $M$ is:
\begin{equation}
    M = \dfrac{\log\left[1-(1-\gamma)^{1/N}\right]}{\log\left(1-\frac{\bar{\alpha}}{N}\right)}
\underset{\bar{\alpha}, \gamma \ll 1 }{\simeq}
 \dfrac{N\log\left(\frac{N}{\gamma}\right)}{\bar{\alpha}}\; .
\end{equation}
For example, with a family-wise error rate $\bar{\alpha}=0.01$, $N=100$ features and $\gamma=0.01$ we can estimate to need at least $M \simeq 9.2 \times 10^4$ samples. Using the binomial distribution, we also get the estimate of $9.2\pm 3.0$ samples in the tail region of each histogram. 

\subsubsection{Moving Histograms}
\label{subsubsec:streaming-histograms}

Once the time steps have been randomly chosen, 
we need to compute the sample histograms, $H_f$, and the corresponding divergence relative to $H_{R,f}$. 

The simplest method to compute a moving histogram, $H_f$, to estimate the distribution of a feature in a given period would be to use a sliding window (e.g., one week) and compute the histogram using all the window events. 
However, in a production environment in \textit{Streaming} this method requires storing and aggregating events in the window, which can be very heavy, especially for long windows and/or use-cases with large event rates.  
Therefore, in our method, we choose to estimate the distribution of features using either an Unbiased Exponential Moving Histogram (UEMH) or its time-based version Unbiased Time-Exponential Moving Histogram (UTEMH) as described in Ref~\cite{menth2017moving}.
Using these methods, no events need to be stored, only the histogram itself at each time step.
The histogram is updated on each incoming event via a recursion formula making the time and memory complexities of this method $O(n\,b)$, with $n$ the number of features and $b$ the number of histogram bins. Since these two quantities are constant and small, we can say that the complexity of the update operation is constant both in time and memory.
All past events contribute to the histogram, $H_f$, but with an exponentially decaying weight, i.e., older events are gradually forgotten.
The half-life, $\tau_{1/2}$ (or the event based version $n_{1/2}$) is the counterpart of the window size (for sliding windows) and controls the timescale of the histogram. It corresponds to the time (or the number of events) until the contribution from a given event is reduced by half.
So, for example, if we want to monitor a timescale of about a week, a half-life of a few days is appropriate to suppress events beyond a week.  
In Fig.~\ref{fig:system-architecture}, we represent this exponential decaying window with a fading dark blue color gradient in the rectangle above the histogram of each sample.

\subsubsection{Distribution of Divergences \& Outputs}

The last piece of the \textit{Training} component inner loop is the computation of the histogram to represent the distribution of divergence values, $H_{D,f}$. This is illustrated in the left bottom green block in Fig.~\ref{fig:system-architecture}, where the arrows connecting to the histogram (\textit{Compare} step) indicate the comparison between the reference histogram $H_{R,f}$ and each sample $H_f$.
Each divergence value contributes to a given bin of $H_{D,f}$.

The divergence measure used in this procedure to compare histograms can be any measure and it does not have to be the same for all features. 
There are numerous measures of divergence available in the literature such as the Kolmogorov-Smirnov (KS), Kuiper, and Anderson-Darling test statistics~\cite{massey1951kolmogorov,kuiper1960tests,anderson1954test} and various information theory divergences such as the Kullback-Leibler (KL) divergence and the Jensen-Shannon Divergence (JSD) (for a recent review see~\cite{cai2020distances}). 
The JSD is well suited to categorical data (though it can also be used for numerical data)
and is well defined even if there are regions where only one distribution has support.
A shortcoming of the JSD is that it has no explicit notion of distance between points in space. Thus, for numerical data we also use the Wasserstein Distance,~\cite{Panaretos_2019}, or the KS measures as alternatives.

Finally, the last step of the \textit{Training} component procedure consists of outputting the final state, for each feature, to be used in the \textit{Monitoring} component. 
%(line~\ref{algo:batch-output} of Algorithm~\ref{batch_algo}).
This is represented schematically in the middle block in Fig.~\ref{fig:system-architecture}.
In summary, for each feature $f$, the initial state for the \textit{Monitoring} component will contain a reference histogram $H_{R,f}$, a histogram of divergence values $H_{D,f}$, and an initial configuration for $H_f$.
The latter is chosen so that the streaming histogram starts in a configuration that represents the reference period to avoid initial artificial alarms (e.g., the last sample in the Training period or a copy of the full reference histogram). 

\subsection{Monitoring Component}
\label{sec:monitoring-component}

After building a reference, the system is ready to monitor a data stream. The right block of Fig.~\ref{fig:system-architecture} contains a diagrammatic representation of the \textit{Monitoring} component of the system, which we detail in the next sections. This is responsible for analyzing periods of data (e.g., one week of data monitored daily) of another dataset (e.g., after the system is deployed in production), to detect the level of divergence in each feature relative to the reference. It can alarm several features, rank them in a scale of severity and output useful visualizations of the evolution of the monitored signals.  

In real-time applications, the monitoring runs over an \emph{unbounded} stream of data, i.e., the streaming computation is supposed to ``run forever''. 
The main monitoring loop of Fig.~\ref{fig:system-architecture} takes the stream 
%($X_{P}$) 
as an input and processes each incoming event, either one by one or periodically.
The input also contains the feature states 
which include, e.g., the reference histograms $H_{R,f}$ and divergence values $H_{D,f}$ per each feature $f$ being monitored. An additional configuration parameter 
specifies the frequency (time or event based) at which the multiple hypothesis testing occurs (e.g., daily). 

\subsubsection{Histogram updates}

The first operation depicted in Fig.~\ref{fig:system-architecture} is the \textit{Update of the Streaming Histograms}.
For a pure streaming implementation this occurs for all incoming events. However, this may also be processed in batches if the monitoring frequency is not event-by-event. For simplicity, we focus on describing the event-by-event update, as indicated by the arrows connecting to the green circle event contributing to the first step of the update. 
This operation has to be performed, for each feature, using the same update method that was used in the \textit{Training} component to build the corresponding samples of streaming histograms.

In our experiments we use UEMA-based histograms.
In the right panel of Fig.~\ref{fig:system-architecture} we observe, in further detail, the steps to update the streaming histograms for each incoming event.
When the latest event arrives, all bin counts in the histogram are suppressed by a common factor according to the type of UEMA histogram used (as discussed in Sect.~\ref{sec:batch-phase}). 
This is either a constant, if event-based, or an exponential of the time difference since the last event, if time-based.
The second step identifies the bin corresponding to the feature value for the incoming event and increases its count by one
(lighter cyan bin increment, pointed to by the arrows - right of Fig.~\ref{fig:system-architecture}).

The histogram update operation is the most computationally demanding component of the system because it is done for each event on the stream. As already discussed in Sect.~\ref{subsubsec:streaming-histograms}, using the Exponential Moving Histograms methods (UEMH and UTEMH), we can reduce the time and space complexity of such an operation to a constant factor that depends only on the number of features under monitoring and the number of histogram bins used.

\subsubsection{Streaming Signals Calculation} 

The next part computes the signals monitored for each feature, to test if an alarm should be raised. 
This is depicted below the histogram updates in Fig.~\ref{fig:system-architecture}.

The process starts with the computation of the divergence between the current streaming histogram, $H_f$, for each feature $f$, and the corresponding reference histogram, $H_{R,f}$. 
This is then located on the $H_{D,f}$ histogram of each feature, represented at the bottom of the diagram in Fig.~\ref{fig:system-architecture}.
The $p$-value for a divergence value to be within the expected distribution of divergences is estimated as:
\begin{equation}
    \text{\emph{p}-value}(d) = 1-CDF_{H_{D,f}}(d)
\end{equation}
where $CDF_{H_{D,f}}$ stands for the Cumulative Distribution Function of $H_{D,f}$ and $d$ is the divergence value observed for feature $f$. 
Each of these $p$-values is represented in the $H_{D,f}$ histograms of Fig.~\ref{fig:system-architecture} by the gray bars to the right of the observed divergence value.

\subsubsection{Multivariate Test}
After all the $p$-values are calculated, a multivariate hypothesis test is applied (see top right box in Fig.~\ref{fig:system-architecture}). We focus on the Holm-Bonferroni correction,~\cite{holm1979simple}, because of its computational simplicity while controlling the family-wise error rate and for not assuming any independence between the tested hypothesis.
The $p$-values are first ordered by ascending value $p_1,\ldots,p_N$.
Note that to each $p$-value $p_i$ we associate a feature $f_i$.
Then we scale each $p$-value $p_i$ to produce a signal, $s_i$, defined as
\begin{equation}
    s_i=p_i \times (N+1-i)\;,\;\; \mathrm{with}\; i= 1,\ldots,N \; .
\end{equation}
Finally, the null hypothesis is rejected if, for any (or several) of the features $f_i$, we have $s_i < \bar{\alpha}$, and an alarm is raised. 
Note that for this test $\bar{\alpha}$ serves as the \textit{threshold} (dashed orange line in Fig.~\ref{fig:system-architecture}).

\subsubsection{Output}

The final step is to generate an explanation to pass to the user of the system, that may help to quickly identify the root cause of the issue.  
 The main elements of the output are:
 \begin{itemize}
      \item \textit{Feature Histograms:} The histograms computed at each step can be used to visualize specific features and their state at a given point in time compared to the corresponding reference histograms.
     \item \textit{$p$-values and signals:} 
    The signals ranking automatically provides a measure of each feature's importance to explain a given divergence, i.e., which features deviate the most from their reference distribution.
    For the Holm-Bonferroni test, this already considers that we are testing several features simultaneously. One useful class of visualizations that efficiently summarize the current and past alarm state of all the available features are heat-maps displaying the signal values of each feature over time, with higher color density for features that are in a stronger alarm state. 
 \end{itemize}
After the output of the system state, the main loop goes back to the beginning, and the system waits for the next event to process.

\section{Experimental Setup}\label{sec:experiments}

In this section we provide a description of the data and of relevant concepts used in the experiments to assess the FM system. 

Our objectives are two-fold. First, we explore the impact of different system configurations on both alarm frequency and duration. Secondly, we aim to determine if the alarms work as intended to properly detect deviating patterns. For the first set of experiments we only use real world production data, which enables stronger production-ready conclusions. In the second set, we also use a publicly available anonymized real world dataset to inject artificial drift. This allows us to control the ground truth drift labels while allowing for the reproducibility of our results.

In both sets of experiments, we build the reference from a training period, and then process the target data to be monitored in a streaming fashion as it would arrive at the system in a production environment. Each monitoring step (computation of histograms, divergences and alarm verification) was performed with a daily frequency.
The frequency of the verification of alarms is chosen considering
the duration of a relevant alarm. 
We define a relevant alarm to last at least the typical debugging time a team of analysts would spend trying to identify the problem. In our use case, we consider daily checks to be a reasonable periodicity.

In the next subsections we describe each of the datasets used, their pre-processing (Sec.~\ref{subsec:data}), and provide details on the hyperparameter space used in the experiments (Sec.~\ref{subsec:configurations}).

\subsection{Data}\label{subsec:data}

We study four different datasets containing credit card transactions, both fraudulent and legitimate. Moreover, these datasets represent two types of entities where financial fraud is present, specifically, merchants and payment processors.

Three of the datasets (two merchants and one payment processor) are private, and, for this reason, the client name and the name of the features in the dataset are anonymized throughout the article. We will refer to these datasets as merchant~1, merchant~2 and payment processor~1. The fourth dataset was made publicly available\footnote{Dataset available at \href{https://www.kaggle.com/c/ieee-fraud-detection/overview}{https://www.kaggle.com/c/ieee-fraud-detection/overview}.} by Vesta Corporation, which is a company providing e-commerce payment solutions. We refer to it as payment processor~2.

The data for merchants~1 and 2 span a period of 22 and 2.5 months, respectively. The data for payment processors~1 and 2 span over 1 year and 6 months, respectively. The daily rate of events per dataset ranges from several thousand to several millions of transactions, with fraud rates at the percent order or smaller. To speed up the experiments with payment processor~1, a random sampling of $2\%$ of the original data was used to reduce its volume to the same order of magnitude as that found in the merchant datasets.

\subsubsection{Features}\label{subsubsec:features}

The features for the proprietary datasets were generated from the raw fields of the transactions. The features of the merchant~1 and merchant~2 datasets were designed by an experienced team of data scientists and fraud analysts. The payment processor~1 dataset had its features generated using an automatic feature engineering system that is focused on designing an extensive set of features to detect fraud patterns~\cite{automl_feedzai_patent}. In both cases, this resulted in several hundreds of features, most of which were numerical and a smaller fraction were categorical. The exception is the public dataset, for which, for the first set of experiments, it was used as it was, with almost all features being categorical.

\subsubsection{Data Splitting}\label{subsubsec:data_splitting}

The 4 datasets were split into reference and streaming sets. For simplicity, we used reference and streaming periods equal to those used for training and testing previous ML models.
For example, for merchant~1, 10 months were used for the reference while the following 12 months were used as the streaming set. Complete information for each dataset is summarized in Table~\ref{tab:summary}.

\begin{table*}[h]
\caption{Summary of each dataset splits, features and average daily transactions.}
\begin{tabular}{@{}llllllll@{}}
\toprule
                    &               & \multicolumn{3}{l}{Features}     & \multicolumn{3}{l}{Timespan (months)} \\ \cmidrule(l){3-8} 
Datasets            & Avg. Daily Tx & Total & Categorical & Numerical & Total    & Reference    & Streaming   \\ \midrule
Merchant 1          & 8.4k          & 183   & 6           & 177       & 22       & 10           & 12          \\
Merchant 2          & 83.3k         & 615   & 183         & 432       & 2.5      & 1.5          & 1           \\
Payment processor 1 & 510.8k        & 272   & 36          & 236       & 12       & 3            & 9           \\
Payment processor 2 (public dataset) & 3.2k          & 363   & 331         & 32        & 6        & 3            & 3           \\ \bottomrule
\end{tabular}
\label{tab:summary}
\end{table*}

\subsection{Parameter Configurations}\label{subsec:configurations}

To achieve the goal of understanding the impact of different system configurations, we varied the following parameters:
\begin{itemize}
    \item \textbf{Half-life} $\tau_{1/2}$: This controls the timescale covered by the target streaming histogram window. The time-based values used were 1 week, 2 weeks and 1 month. This was then converted to an event-based half-life $n_{1/2}$ according to the average event rate.
    \item \textbf{Number of bins} $b$: This parameter sets the number of bins for the histograms. We used three values: $50$, $100$ and $200$.
    \item \textbf{Divergence measure}: We used
    Kolmogorov-Smirnov (KS), Wasserstein (W) and the Jensen-Shannon (JS) for numerical features (as described in Sect.~\ref{sec:methods}, ). For categorical features we always use JS.
\end{itemize}
Finally, we fixed $\gamma$ to 0.01 (see also Sec.~\ref{sec:methods}). Then we ran the FM system for all combinations of the aforementioned parameters. This results in 27 different configurations to experiment for each dataset.

\section{Results}\label{sec:results}

In this section, we present the results obtained after running the FM system for the datasets and scenarios described in Sec.~\ref{sec:experiments}. We start by showing the impact of the various parameters on the multiple runs in Sec.~\ref{subsec:parameter_effects}. Then, in Sec.~\ref{subsec:merchant_signal} and \ref{subsec:public_drift}, we show visualizations that illustrate how the system reacts to feature drifts and help to assess the quality and usefulness of the proposed solution.

\subsection{Effect of the system parameters} \label{subsec:parameter_effects}

To study the sensitivity of the FM system on its configuration parameters, we define two aggregations over the series of alarms obtained in a system run:
\begin{itemize}

    \item \textit{Relative number of chained alarms:} This metric attempts to capture the effective number of distinct alarms within a feature. It is averaged over all features and normalized by the maximum number of chained alarms that could arise in principle, which is equal to a chain of on-off alarm signals.
    
    \item \textit{Average alarm duration:} This is a measure of the average duration of the chained alarms per feature. It is computed as the average of the number of alarms for each chained alarm.
    
\end{itemize}

\begin{figure*}[htbp]
    \centering
    \includegraphics[trim={2cm 0 3cm 0}, clip, width=0.95\textwidth]{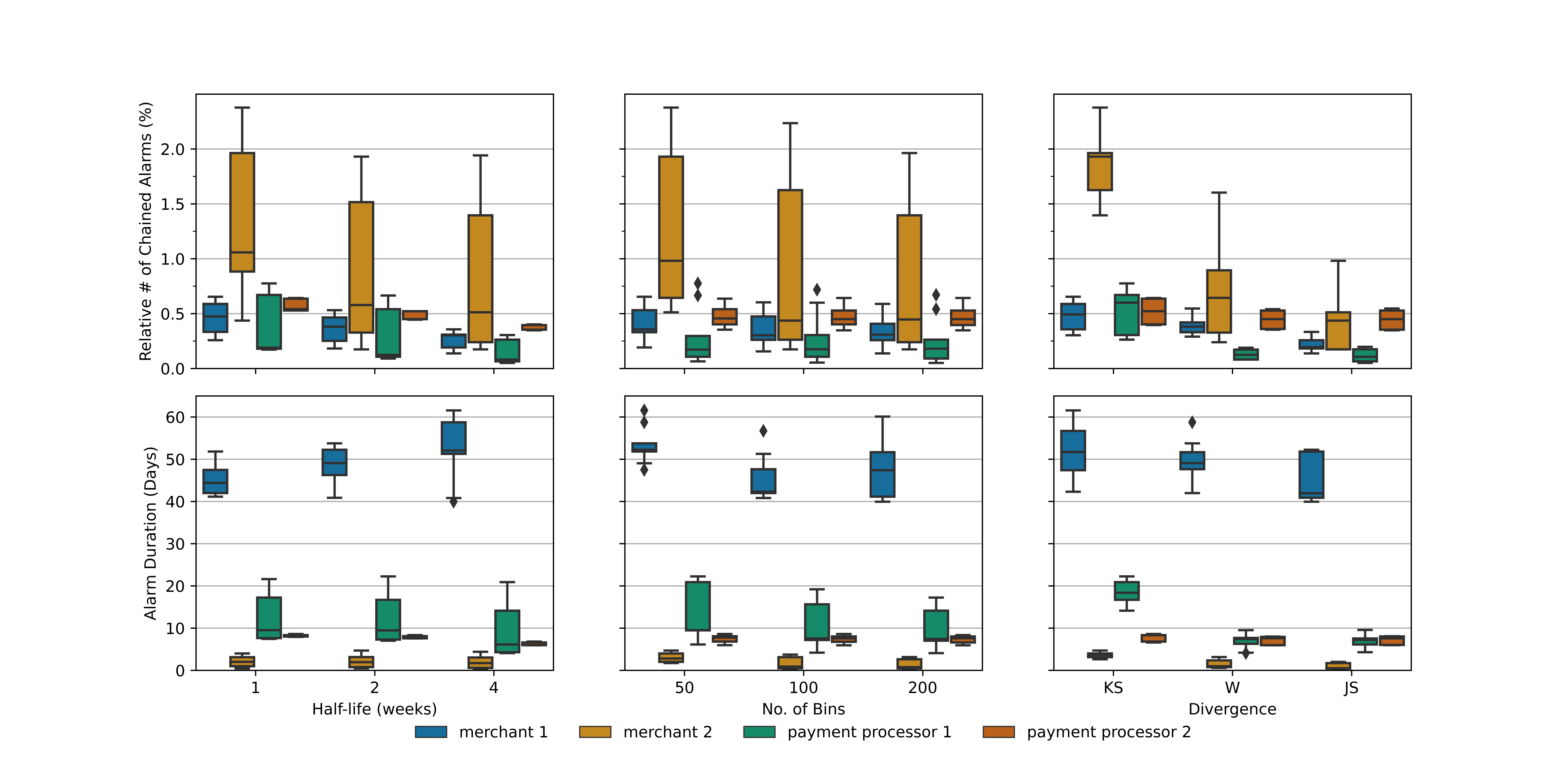}
    \caption{Relative number of chained alarms (top) and average alarm duration (bottom) per feature as a function of the half-life, the number of bins and the divergence measure. The various colors encode the various datasets used.}
    \label{fig:aggregations}
\end{figure*}

In the top row of Fig.~\ref{fig:aggregations} we show the first aggregation, as a function of the various parameters previously specified, for the different datasets. In each box plot, we fix a parameter and represent the distribution of values for all the other combinations of parameters. 

On the left panel, we see that by increasing the half-life, from 1 week to 1 month, the median number of distinct alarms per feature decreases. This is caused by the drop in the sensitivity of the moving histograms to sudden changes of feature values for larger half-lives. The half-life works as a smoothing parameter which controls how much the system reacts to momentary changes of feature values.

In the middle panel, we observe that increasing the number of bins reduces the number of alarmed features. In three of the datasets we see a clear drop in the median from 50 to 100 bins, and then a plateau when increasing to 200. For payment processor~2, however, there is a minimal change in the distribution of values, because the dataset is dominated by categorical features, for which the number of bins is the number of categories.

Finally, on the right panel we observe the effect of using different divergence measures.  Changing from KS to W to JS we observe a decrease in the number of alarms. It is important to note that, for categorical variables we always use JS, since it is the one that is insensitive to ordering, which is usually undefined for categorical variables. This is also the reason for observing little change when switching the divergence measure for payment processor~2.

In the bottom of Fig.~\ref{fig:aggregations} we show analogous plots for the alarm duration. In three of the four datasets we observe a reduction or no change in the median value of the alarm duration as we increase the half-life. Together with the results presented for the first aggregation, we can infer that in these cases the reduction in the system's sensitivity contributes to a lower alarm granularity when compared to smaller half-lives, i.e., shorter drifts are not detected. On the other hand, for merchant~1, we observe an increase of the alarm duration as we increase the half-life. This means that we have a reduced amount of alarms, however larger in size.

Regarding the remaining two parameters, the number of bins and the divergence measure, we observe similar trends to those obtained with the number of chained alarms. Increasing the number of bins (i.e., the resolution) reduces the alarm duration up to the plateau at 100 bins, and changing the divergence measure decreases the alarm duration for the datasets dominated by numerical features.

\subsection{Case Study 1: Merchant Data Anomalies}\label{subsec:merchant_signal}

In this and the next section, we turn to a more granular analysis of the signals and alarms obtained in specific cases. The FM system can only be assessed rigorously in drift cases for which we know the ground truth drift label. These labels are often hard to obtain, since they may rely on human reporting. For this reason we will focus our analysis in cases where we could confirm the existence of a drift in the data. In this section we focus on merchant~1 and two confirmed anomalies found in its data. 
Note that, for brevity, we present the results for a single run of the system with the parameters fixed to 1-week half-life, 100 bins and Wasserstein distance, since the results were qualitatively similar across runs.

\subsubsection{Spike of Card Registrations}
The first issue we discuss was caused by a spike in card registrations in merchant~1. Ideally, problems such as this would be detected by directly monitoring the number of card registrations, however, this feature was not available to be monitored by the system. Nevertheless, the FM system still provided an alarm, as it automatically monitors derived features that are related to card registrations, which served as proxies.
After consulting with the team in charge of this merchant, we were able to trace back this alarm to the spike in card registrations, which the team was unaware of. This is a particularly successful example where the FM system is able to proactively detect problems without the need to manually add specific monitoring metrics to the system.

\begin{figure*}[htbp]
    \centering
    \includegraphics[width=\textwidth]{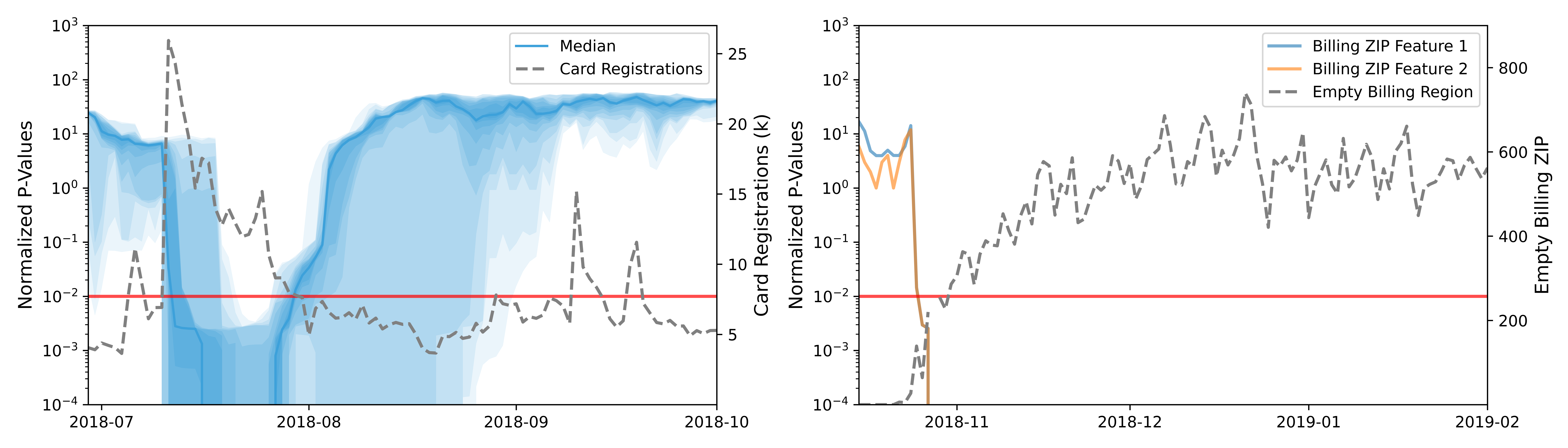}
    \caption{(Left) Distribution of the normalized $p$-values of the features related with the number of card registrations. The median corresponds to the deep blue line and the various shades correspond to 20 evenly spaced quantiles ranging between 0.01 and 0.99. (Right) Normalized $p$-values of the features related with the billing ZIP code, in blue and orange. The alarming threshold corresponds to the red line at the value of 0.01.}
    \label{fig:merchant1_anomalies}
\end{figure*}

In Fig.~\ref{fig:merchant1_anomalies}, on the left, we show the most significant period during which the spike of card registrations took place. The number of card registrations is represented by the dashed gray line and the distribution quantiles of the normalized $p$-values of the features related to the number of card registrations are represented in the various shades of blue, with their median represented by the deeper blue line. We see that the $p$-values start to decrease as the number of card registrations increases and their median keeps decreasing until an alarm is triggered. It then recovers a few days after the peak in card registrations.

\subsubsection{Missing Billing ZIP Codes}
The second issue was caused by an increase in the number of missing billing ZIP codes.
In Fig.~\ref{fig:merchant1_anomalies}, on the right, we observe the normalized $p$-values of two features related to the billing ZIP code (blue and orange lines), which suddenly decrease below the alarm threshold. After exploring the data, we found that there was an increase in the number of missing values in these features during the same period (represented by the dashed gray line). 
This issue was only noticed by the team months after it occurred, whereas the FM system detects it in a few days.

\subsection{Case Study 2: Public Dataset Drift Detection}\label{subsec:public_drift}

The payment processor~2 dataset is publicly available, which allows for our analysis to be reproduced. However, we do not have access to any ground truth information on anomalies or drifts in the data. Therefore, we follow a simulation approach where we inject what could be a common anomaly in this type of data.

We start by defining a perturbation, consisting of transforming some of the transaction amounts from their value in dollars to cents. This could occur by human or system errors. With this goal in mind, we applied the conversion from dollars to cents to $10\%$ of the data, at random, starting 1 month after the beginning of the streaming period, $t0+1M$, and ending by the end of the second month, $t0+2M$.

Each transaction is linked to a single card.
In the dataset the card id is represented by six features, from card1 to card6. We run the automatic feature generator over the transaction amount and card id to create card profiles. These consist of several aggregations used in the classification of fraudulent transactions, such as the average amount or the number of transactions per card in a certain period. In total, we have 26 different features in the final dataset, consisting of the transaction amount and the generated profiles.

\begin{figure}[htbp]
    \centering
    \includegraphics[width=0.49\textwidth]{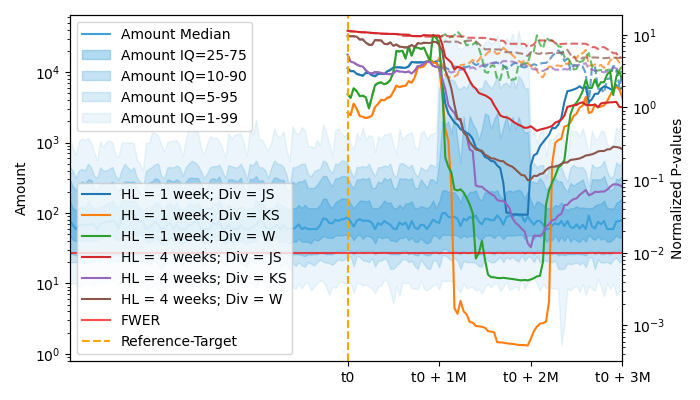}
    \caption{Transaction amount distribution values and normalized $p$-values for various values of the half-life and divergence. The horizontal red line represents the family-wise error rate of 0.01, and, the vertical dashed yellow line marks the transition between the training and monitoring periods.}
    \label{fig:norm_pvalues}
\end{figure}
In Fig.~\ref{fig:norm_pvalues} we represent the distribution of the transaction amount using several quantile bands, computed daily, and illustrated in different tones of blue. The solid coloured lines represent the normalized $p$-values, for that same variable, for a variety of configurations of half-lives and divergence measures, whereas the dashed lines show the normalized $p$-values without anomaly injection.  

We observe that, as intended, the $p$-values start to decrease right after noise injection starts, and conversely, they start to increase as soon as it ends. Comparing the various signals, we can see that the change in the half-life from 1 week to 1 month makes the system react more slowly.
The effect of using different divergence measures is also noticeable. The JS distance has a lower sensitivity to changes in numerical data, while KS is the most sensitive. This higher sensitivity also implies longer alarms, so these parameters should be carefully selected according to the use case.

\begin{figure*}[htbp]
    \centering
    \includegraphics[width=\textwidth]{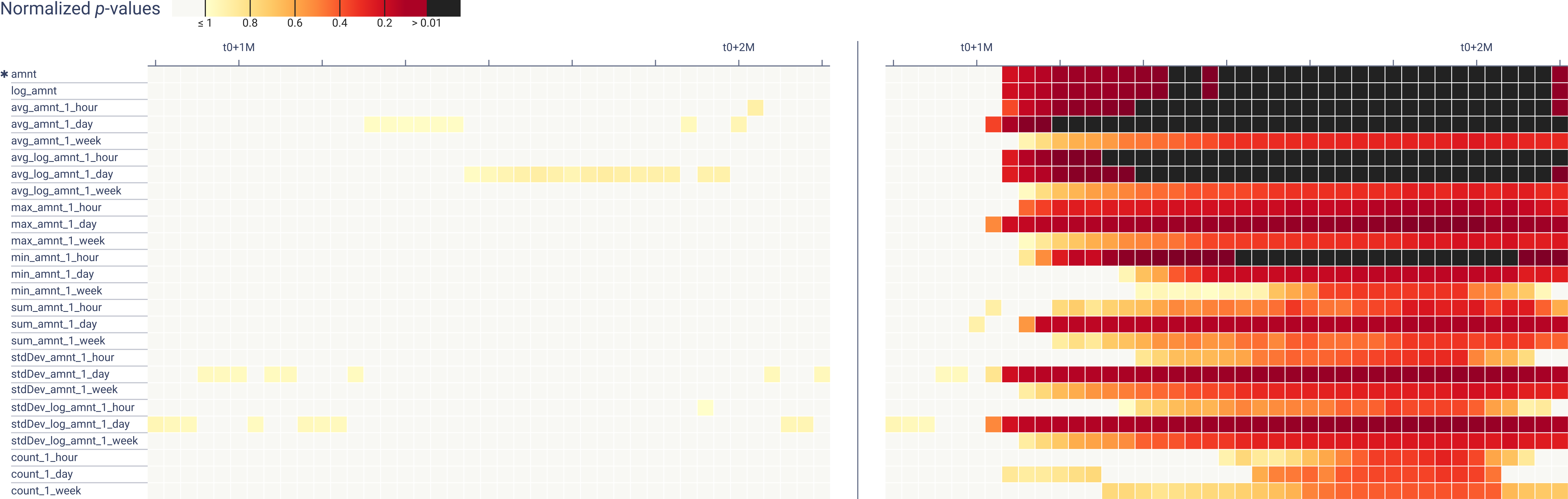}
    \caption{Normalized $p$-values of the various features in the derived dataset before (left) and after (right) injecting noise into the transaction amount. Left key: ``amount'' is abbreviated to ``amnt'' and ``standard deviation'' to ``stdDev''.}
    \label{fig:heatmap}
\end{figure*}
Using the same configuration of the previous section, 1-week half-life, 100 bins and Wasserstein distance, in Fig.~\ref{fig:heatmap} we show two heatmaps representing the normalized $p$-values of the various features around the period of noise injection (see also~\cite{https://doi.org/10.48550/arxiv.2204.14025} for details on this and other visualizations for the system). On the left(right) we represent the normalized $p$-values without(with) the injection of the change. Besides observing a drop in the $p$-values of the amount itself we observe the effect of the change in derived features, such as the average amount or minimum amount. This illustrates how we can identify the problem in the original feature by observing proxy features (with unrelated features presenting the least change).

\section{Related Work}
\label{sec:related-work}

Since in this study we monitor the system in real-time without immediate access to labels we focus this section on unsupervised drift detection methods, i.e., systems that do not distinguish between real or virtual drift (see reference \cite{lu2018learning} for a recent review). 
In this setting, drift detection amounts to an outlier detection problem. Blázquez-García \emph{et al.} reviewed state-of-the-art outlier detection techniques, focusing on time-series data, and presented a taxonomy based on the main aspects of each method, \cite{Blazquez-Garcia-Review-Anomaly-Detection}.
They identify three properties that define the problem and the associated method: 
\begin{enumerate}
    \item \textit{Type of input data}: We focus on multivariate time series.
    \item \textit{Type of outliers we seek to find}: These can be point, subsequence, or time-series outliers, where, respectively, a single point, a portion, or the whole time-series is identified as outlying. We focus on subsequence methods since we monitor a particular period of the series in a sliding manner.
    \item \textit{Type of method to find such outliers}: They can be univariate or multivariate. Univariate methods only use information from the time-series of a single feature to predict its outliers. In contrast, multivariate methods can also use information from several features to predict outliers of a single feature or multiple features. 
    Our system is based on a multivariate method.
\end{enumerate}

The simplest subsequence monitoring approaches in the literature are univariate and single signal. They are typically formulated as a hypothesis test on the distribution of values of a univariate time-series between a reference window and a target window.
Then, the signal to monitor is related to the $p$-value of the observed fluctuation. In Ref.~\cite{kifer2004detecting} two adjacent sliding windows are used to compare reference and target at each time step, with a new set of distribution distance measures providing a significance level as a function of the number of data instances. 
In Ref.~\cite{ross2011nonparametric} a combination of the Mood statistic (to monitor the scale parameter) and the Mann–Whitney statistic (to monitor the location parameter) into a single monitoring statistic is used to avoid the problems of multiple tests. 
In Ref.~\cite{plasse2019multiple} the authors  approximate the multinomial distribution of a stream of values of a categorical feature by using a relative frequencies histogram. 
Their change detection is based on the Kullback Leibler (KL) divergence between static and adaptive estimates of the multinomial densities. They estimate the expected (null hypothesis) distribution of divergences using Monte Carlo simulations. 

Regarding multi-variate methods, most studies focus on monitoring a single signal in order to avoid the problem of multiple hypotheses. 
The \textit{Information-Theoretic Approach (ITA)}, \cite{dasu2006information}, compares two adjacent sliding windows, at each time step using the KL divergence. Relative frequency histograms are used to approximate the distributions for univariate streams and the partitions of a kdqTree are the bins 
for multivariate data streams.
In Ref.~\cite{song2007statistical} (\textit{Statistical Change Detection (SCD)})
a similar windowing approach is used, but a density function is estimated by a multidimensional Kernel Density Estimation (KDE) fitted to the first half of a reference baseline dataset. This is used to compare the log-likelihoods between the target and the second half of the baseline dataset, which is the test statistic. Experiments comparing this method against \cite{dasu2006information} showed a superior statistical power and lower computational execution time.
References~\cite{kuncheva2011change} and~\cite{moshtaghi2016online} leverage Gaussian distribution based representations of the reference data, respectively using a Gaussian Mixture Model and an online multivariate Elliptical Envelope clustering of the data into sets of normal data clusters. In the former, they build a signal from Mahalanobis distances to the test samples, assuming that they follow a chi-square distribution with a number of degrees of freedom given by the dimensionality of the space. For the latter, they build a \emph{state tracker} that aggregates the most recent set of events into another Elliptical Envelope cluster (drift is detected when this is significantly different from existing clusters).
In Ref.~\cite{ding2013anomaly}, the authors propose to use the Isolation Forest model, \cite{liu2008isolation}, to produce scores for each instance in the evaluated stream, that are then aggregated to detect Concept Drift on a set of evaluated instances. 

Other methods that are also multivariate but focus on point outliers, rather than monitoring subsequences,  may, in principle, also be adapted to sub-sequence monitoring,~\cite{yang2009neighbor,tan2011fast,guha2016robust,pham2014anomaly,ebrahimi2018large,pevny2016loda}.

Finally, the set of methods that relate most to our approach are Multivariate Multi-signal based approaches to monitor subsequences. In the \textit{Computational Intelligence-Based CUSUM test (CI-CUSUM)},~\cite{alippi2008just}, the authors propose to extract $12D + \binom{D}{2}$ features from the data, inspired on CUmulative SUM (CUSUM) \cite{page1954continuous} and Mann–Kendall \cite{kendall1948rank} statistics. A Principal Component Analysis (PCA) feature extraction is used to select the set of top-$k$ components from the extracted features. 
In Ref.~\cite{qahtan2015pca} the authors define a method that monitors the top-$k$ components of a PCA transformation of the input space via Kernel Density Estimation. The PCA transformation is obtained in the baseline window. The density functions are obtained for the baseline window and the test window to compute divergence scores. 
Finally, in Ref.~\cite{ahmad2016real}, the authors propose to use Hierarchical Temporal Memory (HTM) as a forecasting model for the dataset evolution. From the forecasts, raw anomaly scores are obtained by comparing the predictions with the actual values of the data points, which are then used to build an anomaly likelihood assuming a normal distribution of raw anomaly scores. 
The system can be broken up into multiple independent HTM models, where the anomaly likelihood is obtained as a joint probability distribution of the raw anomaly scores of each model (assumed to be independent).

\section{Conclusions}\label{sec:conclusions}

In this work we presented a (data driven) lightweight system designed to automatically detect drift in data streams. We started with a detailed formulation of the method, which is based on a strategy of i) first learning estimates of reference distributions for all features, together with representations of their normal levels of fluctuations (\textit{Training} component), and ii) calculating various signals online, comparing them to the reference representations and applying a multivariate statistical test to check for alarms (hence alerting the user for the emergence of an anomaly in the data). We then moved on to an empirical study, using four real world datasets in the domain of credit card fraud detection, where we first presented an analysis of the system's alarming sensitivity as a function of its configurable parameters, followed by illustrative case studies of alarms detected for various datasets.

Regarding the method, its first stage (\textit{Training} component) was designed to be able to tolerate levels of drift as observed in a wide training period, given a shorter monitoring window size scale.
This is achieved by sampling observation windows, which are used to construct reference distributions of divergences, per feature, between the overall training set distributions and the sampled periods. 
In the second stage, a p-value signal per feature is computed while events from the data stream are processed. These are then combined with a multiple hypotheses statistical test, to detect drift alarms together with an explanation via an importance ranking of the alarming features. 
In both stages, feature distributions are represented by UEMA histograms. Thus, in addition to guiding the user towards the important features for the alarm, the method is able to do it with a small constant memory footprint and a small computational cost (via recursive updates).   

As for the experiments, our study of the impact of the various parameters on the multiple system runs, confirmed the intuition that larger half-lives produce less alarms and longer alarms, thus being more appropriate to detect longer term changes. We also observed that a number of bins equal or larger to 100 provides enough resolution to stabilize the patterns of alarms. Finally, we verified that divergence measures that are more suited to numerical features, such as KS or Wasserstein, are more sensitive to changes in such types of features than JS (which is more appropriate for categorical data). In the second part of the experimental study we presented various examples of alarms triggered and concluded that the FM system is able to detect data problems without having to manually add specific metrics to the system to monitor such specific issues (which would be hard to do because we cannot predict in advance all possible issues that might occur in the future). This was both verified for real alarms as well as for synthetically injected anomalies. 

Overall, by applying the feature monitoring system we were able to capture anomalies, some more and others less obvious, that would have otherwise been unnoticed in the data. As expected, as soon as the features start presenting deviating patterns and anomalies we start observing a decrease in their p-values and normalized counterpart, until finally reaching the alarming state, as expected.

Finally, it is important to observe that the triggering of an alarm is related to the value of the FWER, which should be carefully tuned to meet the needs of the application at hand, together with the monitoring half-life. 

Several studies would be interesting to conduct in future work. The use of a sliding reference period (e.g., 3 months with 1 week observation "windows") could bring a different useful view of the system. By applying this approach our method would shift from a static to a dynamic paradigm, an ever evolving system, capable of running indefinitely. There are, however, situations where a static approach presents advantages (or vice versa). In a case where one trains a machine learning model and wants to monitor shifts relative to a training period it may be more advantageous to have a static period. It would also be interesting to perform a more extensive study of the alarming properties of the system by injecting a richer (as well as more numerous) array of anomalies in datasets. In such a study, the goal would be to create a solid ground truth with many alarms with different durations and characteristics. Then one could better study the performance of the system and tune its hyperparameters, e.g., to achieve the best alarm recall given an allowed level of false positive alarms. Another interesting way to assess the performance of the FM system would be in a scenario where the data drift amounts to loss in ML model performance. In such a scenario it would be interesting to understand if the FM system provides good alarms to trigger model retraining and if it would be more effective than periodically retraining the system (thus saving unnecessary ML model re-training actions).

\begin{acks}
We thank Rita Costa, Beatriz Jorge, João Palmeiro and David Polido, from Feedzai's Data Visualization team, for making the visualization system into reality as well as for feedback on the content of the manuscript. We also thank Ricardo Barata, João Oliveirinha, João Veiga and Sofia Gomes for discussions during the project.\vspace{-3pt}
\end{acks}

% %%
% %% If your work has an appendix, this is the place to put it.
%\appendix
%\input{supplementary_material}

%\clearpage
\bibliographystyle{ACM-Reference-Format}
\bibliography{references}
% \clearpage

\end{document}